\documentclass[letterpaper, 10 pt, conference]{ieeeconf}  

\IEEEoverridecommandlockouts                              

\usepackage{cite}
\usepackage[table]{xcolor}
\usepackage{graphicx}
\usepackage{array,booktabs,arydshln}
\usepackage{hyperref}
\usepackage{algorithm}
\usepackage{algpseudocode}

\definecolor{green}{RGB}{3,112,15}

\title{\LARGE \bf
Benchmarking Metric Ground Navigation
}

\author{Daniel Perille$^{1}$, Abigail Truong$^{1}$, Xuesu Xiao$^{1}$, and Peter Stone$^{1}$
\thanks{$^{1}$Daniel Perille,  Abigail Truong, Xuesu Xiao, and Peter Stone are with Department of Computer Science, University of Texas at Austin, Austin, TX 78712 {\tt\scriptsize \{danny.perille, a.truong\}@utexas.edu}, {\tt\scriptsize \{xiao, pstone\}@cs.utexas.edu}}
}

\begin{document}

\maketitle
\thispagestyle{empty}
\pagestyle{empty}

\begin{abstract}
Metric ground navigation addresses the problem of autonomously moving a robot from one point to another in an obstacle-occupied planar environment in a collision-free manner. It is one of the most fundamental capabilities of intelligent mobile robots. 
This paper presents a standardized testbed with a set of environments and metrics to benchmark difficulty of different scenarios and performance of different systems of metric ground navigation. 
Current benchmarks focus on individual components of mobile robot navigation, such as perception and state estimation, but the navigation performance as a whole is rarely measured in a systematic and standardized fashion. 
As a result, navigation systems are usually tested and compared in an ad hoc manner, such as in one or two manually chosen environments. 
The introduced benchmark provides a general testbed for ground robot navigation in a metric world. The Benchmark for Autonomous Robot Navigation (BARN) dataset includes 300 navigation environments, which are ordered by a set of difficulty metrics. Navigation performance can be tested and compared in those environments in a systematic and objective fashion. 
This benchmark can be used to predict navigation difficulty of a new environment, compare navigation systems, and potentially serve as a cost function and a curriculum for planning-based and learning-based navigation systems. We have published our dataset and the source code to generate datasets for different robot footprints at \url{www.cs.utexas.edu/~xiao/BARN/BARN.html}.
\end{abstract}

\section{INTRODUCTION}
\label{sec::introduction}

Autonomously moving from one point to another, especially in challenging environments, is one essential capability of intelligent mobile robots. This problem of mobile robot navigation has been studied by the robotics community for decades~\cite{quinlan1993elastic, fox1997dynamic, knotts1998navigates}. Sophisticated navigation systems have been developed using classical control methods~\cite{knotts1998navigates, xiao2017uav, xiao2015locomotive}, path and motion planning~\cite{quinlan1993elastic, fox1997dynamic, xiao2018motion}, or, more recently, machine learning techniques~\cite{giusti2015machine, pfeiffer2017perception, xiao2020appld, xiao2020toward, liu2020lifelong}. 

However, despite the plethora of works in mobile robot navigation, there is no generally accepted metric by which to compare different approaches against one another, even for navigation in a simple metric world, where only geometric obstacles are considered. Although in relatively open space, navigation performance of different systems may not vary significantly, the lack of an accepted metric becomes particularly relevant in environments that are more difficult to navigate. Those environments include, for example, unstructured or confined spaces for search and rescue~\cite{xiao2020risk} and highly constrained spaces where agile maneuvers are required for robots with nonholonomic motion constraints~\cite{xiao2020toward}. Newly developed navigation systems are only tested and compared to existing ones in a limited number of ad hoc environments with unquantified difficulties. 

To address the lack of a standardized method to test and compare mobile robot navigation systems, this work provides a Benchmark for Autonomous Robot Navigation (BARN) of 300 simulated test environments, which can also be easily instantiated in the physical world. We design a set of metrics to quantify navigation difficulty of these simulated environments for ground mobile robots to move in obstacle-occupied spaces without collision. 
We then unify this set of metrics via a learned function approximator to introduce a novel measure of an environment's difficulty level for metric ground navigation.
Through an extensive amount of 3000 simulated trials using two widely used planners, Dynamic Window Approach (DWA)~\cite{fox1997dynamic} and Elastic Bands (E-Band)~\cite{quinlan1993elastic}, we benchmark the relative difficulty levels for the test environments. 
We also run multiple physical navigation trials to validate our model's predictive power of navigation difficulty. 
To summarize, the main contributions of this work are: 

\begin{itemize}
    \item A benchmark dataset (BARN) of 300 simulated test environments for metric ground navigation, 
    \item A set of metrics and a data-driven model to combine them to quantify the challenge posed by a particular environment for mobile robot navigation.
\end{itemize}

The rest of the paper is organized as follows: Section \ref{sec::related} reviews existing benchmarks related to mobile robot navigation. Section \ref{sec::approach} describes our method of constructing BARN, including the test environments, the set of difficulty metrics, and the data-driven model to combine them. Section \ref{sec::experiments} provides implementation details and experiment results to validate that the proposed difficulty metrics and the learned function approximator can predict navigation difficulties using a physical mobile robot in the real world. 
Section \ref{sec::conclusions} concludes the paper. 
\section{RELATED WORK}
\label{sec::related}
This section reviews existing testbeds and metrics related to mobile robot navigation. 

\subsection{Testbeds}
Testbeds are designed as an apparatus to quantify performance based on a set of pre-defined metrics. The tests can be replicated when following a standardized testing procedure. 

\subsubsection{Physical Testbeds}
National Institute of Standards and Technology (NIST) has created standard testbeds for response robots, including ground, aerial, and aquatic vehicles~\cite{NIST}. Specialized test methods are developed to test individual robot capacity, e.g. locomotion, sensing, communication, durability. Robotarium~\cite{pickem2017robotarium} is a testbed developed to test algorithms for multi-robot research, using a fleet of miniature differential drive robots. Xiao et al~\cite{xiao2018review} reviewed 20 physical testbeds for snake robots and pointed out that all of them are designed in an ad hoc manner, i.e. tailored to demonstrate the newly developed capability. They also provided suggestions on a general testbed design. The research thrust on developing robot testbeds demonstrates the robotics community's need for standardized test methods to quantify robot performance and research progress. Similar to the aforementioned testbeds but with a different purpose, the proposed testbed is developed to benchmark mobile robot navigation systems operating in a metric world. 

\subsubsection{Software Testbeds}
Thanks to the recent progress on data-driven approaches, testbeds are frequently  instantiated as datasets, e.g. ImageNet~\cite{deng2009imagenet}. In the mobile robot navigation domain, especially on the perception and estimation side, many such software testbeds have also been created~\cite{sturm2012benchmark, geiger2013vision, maddern20171, tartanair2020iros}, where perceptual data is collected along a fixed motion trajectory. However, when arbitrary motion execution is required, such interactive testbeds become sparse. Even when motion is allowed \cite{kolve2017ai2, savva2019habitat}, the locomotion part of navigation is assumed to be trivial, i.e. the testbeds only benchmark the robot's ability to infer ``where'' to navigate, instead of to generate feasible and optimal motion commands for ``how'' to navigate. The proposed testbed focuses on the ability to autonomously generate viable motion commands in order to navigate between two fixed points in a given environment. Since only geometric obstacles are considered, it can easily be instantiated into a physical testbed. 

\subsection{Metrics}
A common metric to quantify mobile robot navigation difficulty is distance from points on the path to the closest obstacle \cite{de2011minimum, feyzabadi2014risk}, as the closer the robot needs to come to an obstacle, the more difficult the navigation task. Past experiences (e.g. previous failure cases) have also been utilized to quantify difficulty as a function of a single state when navigating in the ocean \cite{pereira2013risk} or in city traffic \cite{krumm2017risk}. Not many works considered more than one single source of difficulty: Soltani et al. \cite{soltani2004fuzzy} represented difficulty with both distance to closest obstacle and visibility of a particular location and combined their effects using manually defined weights. Robot motion risk can also be viewed as an indication of difficulty: recent risk reasoning frameworks extended the dependency of risk associated with a certain state into motion history \cite{xiao2020robot, xiao2020risk}, and pointed out that difficulty/risk caused by, for example, turning or dragging a tether, cannot be determined by a single state alone. The multiple difficulty metrics designed in this work are inspired by the risk universe \cite{xiao2020risk}. In order to determine the combined effect of all individual elements, we use a data-driven approach instead of manually assigned weights. 

\section{APPROACH}
\label{sec::approach}
In this section, we describe our method of constructing BARN. The navigation environments are first generated using cellular automaton \cite{wolfram1983statistical}, for which navigational paths are planned on the robot Configuration Space (C-Space) \cite{latombe2012robot}. 
Second, we introduce a set of metrics used to quantify navigation difficulty level. Third, a function approximator is learned in a data-driven manner to combine these difficulty metrics and determine the final difficulty level of navigating through a specific environment. 

\subsection{Navigation Environments}
\begin{figure}[b]
\centering
\includegraphics[width=1\columnwidth]{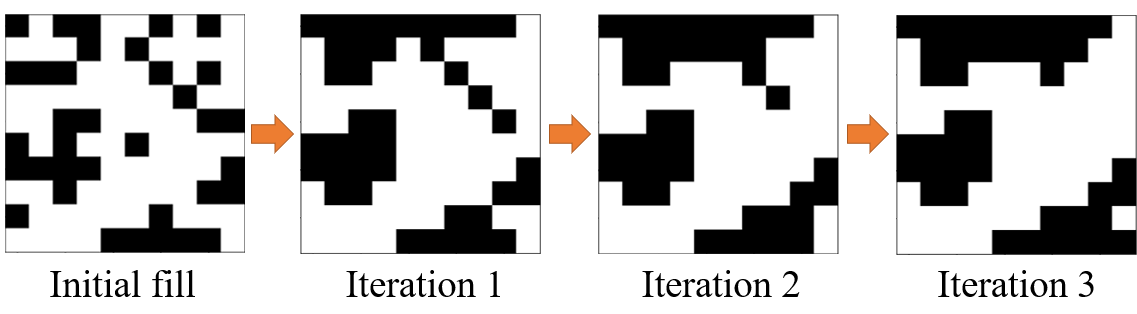}
\vspace{-5pt}
\caption{Three \emph{smoothing iterations} of a cellular automaton with an \emph{initial fill percentage} of 0.35, \emph{fill threshold} of 5, and \emph{clear threshold} of 1}
\label{fig::cellular_automata_figure}
\end{figure}

Navigation environments are systematically generated through the method of cellular automaton~\cite{wolfram1983statistical}. A cellular automaton was originally designed as a collection of black cells on a white grid of specified shape that evolves through a number of discrete time steps according to a set of rules based on the states of neighboring cells~\cite{wolfram1983statistical}. In this work, we use black cells to represent obstacle-occupied space and white cells to represent free space. The evolution of the black cells generates different obstacle configurations. Cellular automaton is easy to scale to any size, generates more realistic environments than random fill, and is also easily customizable due to its parameters that can be changed to generate different types of worlds. Due to the smoothing iterations, the resulting grid resembles real-world obstacles more than the initial randomly filled grid does. We use four parameters of the cellular automaton to control the generation of obstacles: \emph{initial fill percentage}, \emph{smoothing iterations}, \emph{fill threshold}, and \emph{clear threshold}. The procedure to generate navigation environments using cellular automaton is provided in Algorithm \ref{alg::cellular_automaton}, with an example in Figure \ref{fig::cellular_automata_figure}. 

        
            
\begin{algorithm}[ht]
    \caption{Navigation Environments Generation}
    \label{alg::cellular_automaton}
    \begin{algorithmic}[1]
    \State \textbf{Inputs}: $m$, $n$, \emph{initial fill percentage}, \emph{smoothing iterations}, \emph{fill threshold}, \emph{clear threshold},
    \State Randomly fill a $m\times n$ grid of $0$'s with \emph{initial fill percentage} of $1$'s
    \For{iteration $k = 1 : \emph{smoothing iterations}$}
        \For{cell in grid}
            \If{$|$FilledNeighbors(cell)$|$ $\geq$ \emph{fill threshold}}
                \State cell $\leftarrow 1$ \algorithmiccomment{Fill cell}
            \EndIf
            \If{$|$FilledNeighbors(cell)$|$ $\leq$ \emph{clear threshold}}
                \State cell $\leftarrow$ 0 \algorithmiccomment{Empty cell}
            \EndIf
        \EndFor
    \EndFor
    \end{algorithmic}
\end{algorithm}

Each obstacle grid is then converted into the robot's C-space based on the robot dimension. In the C-space, one free point on both the left and right edge are chosen at random to be the start and end points of the path, respectively. A flood-fill algorithm~\cite{torbert2016applied} is used to determine if there is an open path between the points. If no path is possible, then the space is discarded. A* algorithm~\cite{Hart1968} is then used to plan a path in the free C-space. 

\subsection{Difficulty Metrics}
Upon generating an environment through cellular automaton and establishing a path therein, various metrics are calculated in the C-space along the path to quantify the difficulty of traversal.

\subsubsection{Distance to Closest Obstacle}
At each cell in the environment, the Distance to Closest Obstacle is defined as the distance from this cell to the nearest occupied space. This metric is averaged over all points in the path.

\subsubsection{Average Visibility}
\begin{figure}[ht]
\centering
\includegraphics[width=0.8\columnwidth]{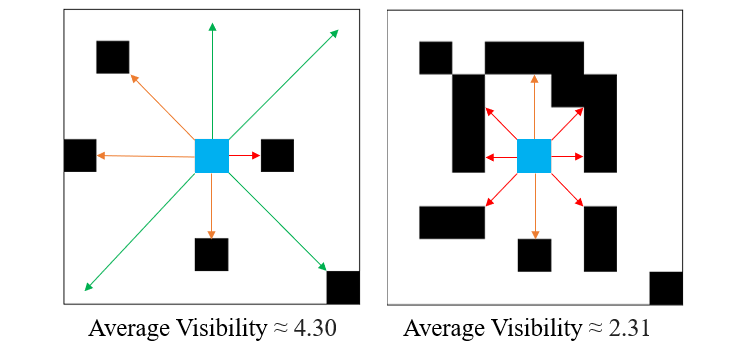}
\caption{An example of high and low Average Visibility}
\label{fig::visibility_figure}
\vspace{-5pt}
\end{figure}

A cell's Average Visibility is defined as the average of the distances to an obstacle along each ray in a $360^{\circ}$ scan. In our discrete space, we average the visibility along eight rays (four cardinal directions and four diagonals), then average this metric over all points in the path. Figure \ref{fig::visibility_figure} provides examples of high and low visibility.

\subsubsection{Dispersion}
\begin{figure}[ht]
\centering
\includegraphics[width=0.9\columnwidth]{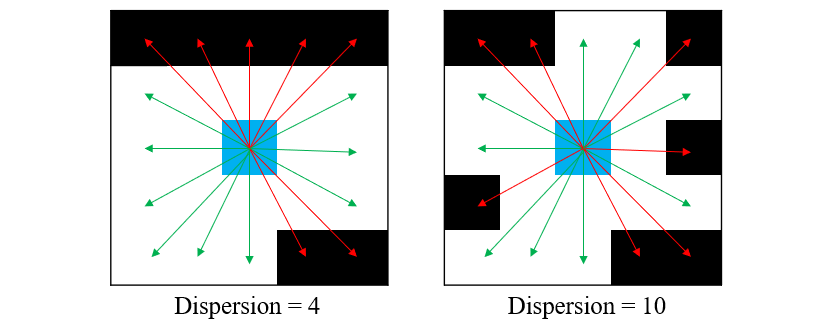}
\caption{Dispersion represents the alternations from occupied to unoccupied space or vice versa. }
\label{fig::dispersion_figure}
\vspace{-5pt}
\end{figure}

From a given state in the environment, a $360^{\circ}$ scan is cast up to a certain max length. The dispersion at that state is defined as the number of alternations in that scan from occupied to unoccupied space or vice versa, as shown in Figure \ref{fig::dispersion_figure}. 

Dispersion captures the number of potential paths out of a given location. A higher dispersion means more possible options for the navigation algorithm and therefore means the environment poses more challenges, especially for a sampling-based local planner, like DWA~\cite{fox1997dynamic}.
In our discrete space, dispersion is calculated by casting 16 rays up to a max length, and checking which are blocked or open. This metric is calculated for each point in the path and averaged over the length of the path.

\subsubsection{Characteristic Dimension}
\begin{figure}[ht]
\centering
\includegraphics[width=0.8\columnwidth]{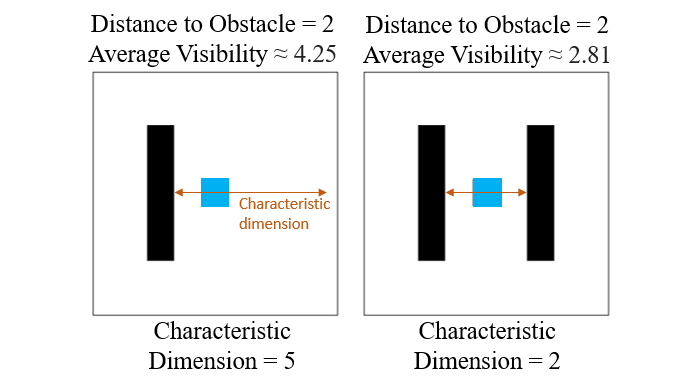}
\caption{Characteristic Dimension captures the tightness of a state in an environment. }
\label{fig::char_dim_figure}
\vspace{-5pt}
\end{figure}

The Characteristic Dimension at a given cell is defined as the visibility of the axis through the cell with the lowest visibility. 
In our discrete space, 8 axes (each made up of 2 rays $180 ^{\circ}$ apart) are cast $22.5 ^{\circ}$ apart from one another. Each axis' visibility is calculated as the sum of the distances to an obstacle along both of the rays that make it up. The Characteristic Dimension is defined as the visibility of the axis with the lowest visibility.
The Characteristic Dimension captures the tightness of a space. A low Distance to Closest Obstacle could occur in a relatively open space and therefore still be quite easy to navigate. Additionally, a space might be tight along one axis yet very open along another (e.g. a long tunnel) and therefore have a high Average Visibility despite being narrow. Figure \ref{fig::char_dim_figure} captures such an instance of when Distance to Closest Obstacle and Average Visibility may fail to completely represent the difficulty of a space.

\subsubsection{Tortuosity}
Tortuosity, as a property of a curve being tortuous, is calculated over the entire path using the arc-chord ratio. The arc length is the length of the entire path, while the chord length is the length of a straight line between the start and end points. This metric captures bends in the path that make navigation more difficult compared to navigation along a relatively straight path.

\subsection{Combined Difficulty Level}
We first use a data-driven approach to benchmark the relative difficulty level of all the navigation environments in our dataset. Thousands of simulation trials using representative and widely-used navigation systems are conducted to reveal the actual difficulty of an environment. Second, to investigate how the difficulty metrics interact with each other and contribute to the combined difficulty measure, we learn a function approximator to map from the individual difficulty metrics to the final difficulty level of a given environment. This model can be used to predict difficulty of unseen navigation environments. Please refer to Section \ref{sec::experiments} for details. 

\section{EXPERIMENTS}
\label{sec::experiments}
In this section, we present implementation details of our dataset generation and physical experiments using a ground robot to validate that our benchmark model can accurately predict the difficulty level of unseen physical navigation environments. 

\subsection{Dataset Generation}

We use 12 sets of cellular automaton parameters to generate the 300 navigation environments of 30 by 30 cells in our dataset, shown in Table \ref{tab::cellular_automaton_parameters}. The parameters are chosen to generate varied worlds with reasonably realistic configurations. Each of those parameter sets is repeated 25 times. We use a Clearpath Jackal robot's dimension (0.508m by 0.430m, corresponding to 5 by 5 cells) to inflate the obstacles and generate the C-spaces. We also provide the original obstacle map in the dataset so that C-spaces corresponding to different robot sizes can be generated by the users.



\begin{table}[ht]
\centering
\caption{Cellular Automaton Parameters}
\begin{tabular}{cccc}
\toprule
Parameters & Values \\
\midrule
\emph{initial fill percentage} & \{0.15, 0.20, 0.25, 0.30\}\\
\emph{smoothing iterations}    & \{2, 3, 4\}  \\
\emph{fill threshold}          & 5 \\
\emph{clear threshold}         & 1  \\
\emph{neighborhood}          & 8 \\
\midrule
\textbf{Repetitions}    & 25 \\
\bottomrule
\end{tabular}
\label{tab::cellular_automaton_parameters}
\end{table}

The minimum, maximum, mean, and standard deviation of the five difficulty metrics present in the dataset are shown in Table \ref{tab::min_max}. For a given cell, the minimum Distance to Closest Obstacle is 1 for an unoccupied space, and the maximum is 10.20. 
We cast eight rays to compute Average Visibility, resulting in a range from 1 to 14. 
The minimum dispersion is 0 (no alternations between occupied and unoccupied spaces along the 16 axes) and the maximum is 12. 
The minimum Characteristic Dimension is 0 (completely closed space) and the maximum is 20.00. The four metrics above are computed for and averaged over all states on a path. 
The minimum tortuosity of a path is 1 (straight line) and the maximum is 1.71. 

\begin{table}[ht]
\centering
\caption{Difficulty Metric Values}
\begin{tabular}{ccccc}
\toprule
                        & Min.    & Max.  & Mean & Std. \\
\midrule
Distance to Closest Obstacle & 1 & 10.20  & 2.37 &  0.93\\
Average Visibility    & 1   & 14.00  &  4.42 &  1.64\\
Dispersion         & 0 & 12 &  4.35 &  0.89 \\
Characteristic Dimension         & 0    & 20.00  &  4.05 &  2.66 \\
\hdashline
Tortuosity       & 1    & 1.71  &  1.21 &  0.14 \\
\bottomrule
\end{tabular}
\label{tab::min_max}
\end{table}



\subsubsection{Simulation Trials}
We use a simulated Clearpath Jackal, a four-wheeled, differential drive, nonholonomic ground robot, in a Robot Operating System (ROS) Gazebo simulator \cite{koenig2004design} to benchmark the relative difficulty levels of the navigation environments in our dataset. 

We choose two different widely used navigation planners, DWA \cite{fox1997dynamic} and E-Band \cite{quinlan1993elastic} to navigate Jackal. DWA is a representative sampling-based motion planner. Given a global path produced by the A* 
algorithm \cite{Hart1968}, DWA generates samples of linear and angular velocities and evaluates the score of each sample based on closeness to the obstacle, to the global path, and progress toward the local goal. The action sample with the best score is executed to move the robot. The randomness in the sampling process leads to non-deterministic behavior in the same environment. E-Band is a representative optimization-based motion planner, which optimizes an initial trajectory. It deforms the trajectory using virtual bubbles along it, which are subject to repulsive force from the obstacles. The optimized trajectory acts like an elastic band. The default navigation planner from the robot manufacturer, Clearpath Robotics, is the DWA planner. We use the default planner parameters from the manufacturer\footnote{\url{https://github.com/jackal/jackal/tree/melodic-devel/jackal_navigation/params}} for DWA, and the default parameters from the E-Band designer\footnote{\url{http://wiki.ros.org/eband_local_planner}} for the E-Band planner. We set E-Band's maximum allowable linear and angular velocities (0.5m/s and 1.57rad/s) to match with those of DWA for a fair comparison. 

\begin{figure*}[ht]
    \centering
    \includegraphics[width=2\columnwidth]{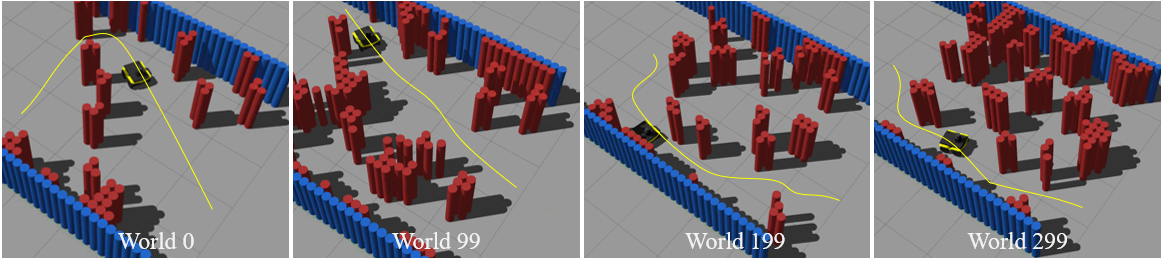}
    \caption{Four Example Environments in Gazebo Simulation (ordered by ascending relative difficulty level)}
    \label{fig::gazebo_images}
    \vspace{-10pt}
\end{figure*}

\begin{figure}[ht!]
    \centering
    \includegraphics[width=1\columnwidth]{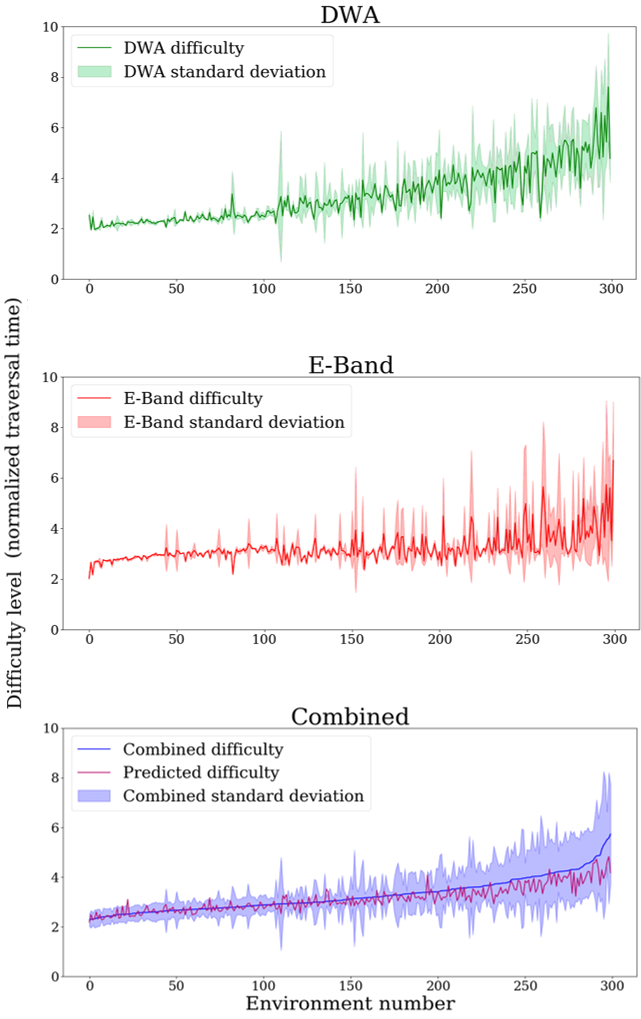}
    \vspace{-15pt}
    \caption{Environment Difficulty Benchmarked by Navigation Performance of DWA and E-Band}
    \label{fig::planner_plots}
    \vspace{-12pt}
\end{figure}

For each one of the 300 environments in our dataset, a pre-built map is provided to the planner, and we run five trials each for DWA and E-Band, resulting in a total number of 3000 trials (Figure \ref{fig::gazebo_images}). The final difficulty measure is the traversal time averaged over the ten trials and normalized by path length. We also compute the variance of the traversal time. A 30-second penalty is introduced to trials where the robot fails to reach the goal, e.g. getting stuck. The DWA, E-Band, and combined (averaged) with predicted navigation performance is shown in Figure \ref{fig::planner_plots}. From left to right, the 300 environments are ordered from easy to difficult. High difficulty level is correlated with high variance. Operating with a map, DWA and E-Band achieve an average normalized traversal time of $3.30\pm0.50$s/m and $3.24\pm0.38$s/m, respectively. As a sampling-based planner, DWA results in higher standard deviation than the optimization-based E-Band, and is more sensitive to the increased difficulty level.  

\begin{figure*}[ht]
\centering
\includegraphics[width=2\columnwidth]{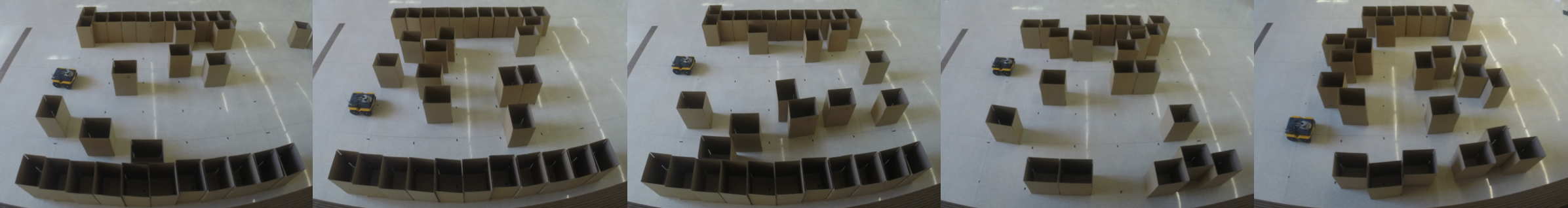}
\caption{For each of the five physical navigation environments, five DWA and five E-Band trials are conducted.}
\label{fig::physical}
\vspace{-3pt}
\end{figure*}

\subsubsection{Function Approximator}
To approximate the combined effect of all five difficulty metrics, we use a simple neural network consisting of two fully connected layers with 64 neurons each. Distance to Closest Obstacle, Average Visibility, Dispersion, and Characteristic Dimension are computed for and averaged over all the states on a path. Tortuosity is computed for the entire path. All these metrics are normalized based on their mean and standard deviation (Table \ref{tab::min_max}). The label of the neural network output is the average traversal time normalized by path length, 
computed from the 3000 simulation trials. 
Our function approximator can achieve a 0.10 prediction loss on normalized traversal time, which corresponds to, for example, 0.50 seconds error while traversing a 5m long path. 

\subsection{Physical Experiments}
\begin{figure}[ht]
\centering
\includegraphics[width=0.8\columnwidth]{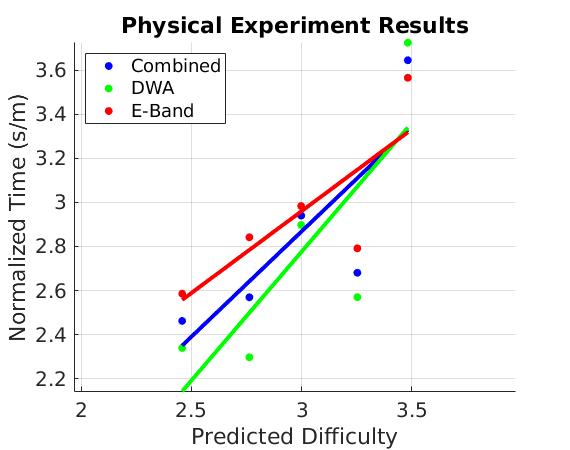}
\caption{Real-World Navigation Performance vs. Predicted Difficulty}
\label{fig::physical_results}
\vspace{-5pt}
\end{figure}

To validate our benchmarks in the real world, we also conduct physical trials in unseen navigation environments. We use a physical Jackal with DWA and E-Band planners parameterized in the same way as in the simulation trials. Five new navigation environments are created using cellular automaton and instantiated in the real-world with cardboard boxes representing obstacles (Figure \ref{fig::physical}). We run five trials with each planner in each environment, resulting in a total of 50 physical trials. Unlike the simulated trials, the planners do not have access to a pre-built map. As shown in Figure \ref{fig::physical_results}, higher predicted difficulty corresponds to longer normalized traversal time (The fitted blue line has a slope of 0.96 and almost zero intercept). In physical environments with low difficulty, DWA performs better than E-Band without a pre-built map. However, the steeper slope of the green line (1.17) than that of the red line (0.74) indicates that DWA is more sensitive to increased difficulty level than E-Band is, which is a similar trend we observe in simulation.

\section{CONCLUSIONS}
\label{sec::conclusions}
We present a dataset\footnote{\url{www.cs.utexas.edu/~attruong/metrics_dataset.html}} of 300 simulated navigation environments and five difficulty metrics along with a data-driven model to quantify the difficulty measure of a particular environment for mobile robot navigation. 
We benchmark the relative difficulty level using 3000 simulated navigation trials with two widely used navigation planners, DWA and E-Band, which are representative of sampling-based and optimization-based planners, respectively. Our model can predict the difficulty of unseen navigation environments based on the five difficulty metrics, i.e. Distance to Closest Obstacle, Average Visibility, Dispersion, Characteristic Dimension, and Tortuosity. 50 physical experiment trials demonstrate that the difficulty level predicted by our model corresponds to real-world performance in unseen environments. As a general testbed, metric ground navigation performance of different systems can be tested and compared with each other in a systematic and objective fashion. The difficulty metrics and the learned function approximator can be used as a new cost function and a curriculum for planning-based and learning-based navigation systems. 







\section*{ACKNOWLEDGMENT}

This work has taken place in the Learning Agents Research
Group (LARG) at UT Austin.  LARG research is supported in part by NSF
(CPS-1739964, IIS-1724157, NRI-1925082), ONR (N00014-18-2243), FLI
(RFP2-000), ARO (W911NF-19-2-0333), DARPA, Lockheed Martin, GM, and
Bosch.  Peter Stone serves as the Executive Director of Sony AI
America and receives financial compensation for this work.  The terms
of this arrangement have been reviewed and approved by the University
of Texas at Austin in accordance with its policy on objectivity in
research.


\bibliographystyle{IEEEtran}
\bibliography{IEEEabrv,references}

\end{document}